\documentclass[letterpaper]{article} % DO NOT CHANGE THIS
\usepackage{aaai2026}  % DO NOT CHANGE THIS
\usepackage{times}  % DO NOT CHANGE THIS
\usepackage{helvet}  % DO NOT CHANGE THIS
\usepackage{courier}  % DO NOT CHANGE THIS
\usepackage[hyphens]{url}  % DO NOT CHANGE THIS
\usepackage{graphicx} % DO NOT CHANGE THIS
\usepackage{natbib}  % DO NOT CHANGE THIS AND DO NOT ADD ANY OPTIONS TO IT
\usepackage{caption} % DO NOT CHANGE THIS AND DO NOT ADD ANY OPTIONS TO IT
\usepackage{algorithm}
\usepackage{algorithmic}
\usepackage{bm}
\usepackage{amsmath, amsfonts}
\usepackage{booktabs}
\usepackage{multirow}

\usepackage{newfloat}
\usepackage{listings}
\DeclareCaptionStyle{ruled}{labelfont=normalfont,labelsep=colon,strut=off} % DO NOT CHANGE THIS
\floatstyle{ruled}
\newfloat{listing}{tb}{lst}{}
\floatname{listing}{Listing}
\title{Towards Adaptive Humanoid Control via Multi-Behavior Distillation \\and Reinforced Fine-Tuning}
\author{
    Yingnan Zhao\textsuperscript{\rm 1,7},
    Xinmiao Wang\textsuperscript{\rm 1,2},
    Dewei Wang\textsuperscript{\rm 2,3}\footnotemark[2],
    Xinzhe Liu\textsuperscript{\rm 2,4},
    Dan Lu\textsuperscript{\rm 1,7}\footnotemark[1], \\ 
    Qilong Han\textsuperscript{\rm 1,7},
    Peng Liu\textsuperscript{\rm 5},
    Chenjia Bai\textsuperscript{\rm 2,6}\footnotemark[1] 
}

\affiliations{
    \textsuperscript{\rm 1}College of Computer Science and Technology, Harbin Engineering University\\
    \textsuperscript{\rm 2}Institute of Artificial Intelligence (TeleAI), China Telecom\\
    \textsuperscript{\rm 3}School of Information Science and Technology, University of Science and Technology of China\\
    \textsuperscript{\rm 4}School of Information Science and Technology, ShanghaiTech University\\
    \textsuperscript{\rm 5}College of Computer Science and Technology, Harbin Institute of Technology\\
    \textsuperscript{\rm 6}Shenzhen Research Institute of Northwestern Polytechnical University \\
    \textsuperscript{\rm 7}National Engineering Laboratory for Modeling and Emulation in E-Government, Harbin Engineering University\\

    \{zhaoyingnan, wangxinmiao\}@hrbeu.edu.cn, bz24006012@mail.ustc.edu.cn, liuxzh12023@shanghaitech.edu.cn, \\
    \{ludan, hanqilong\}@hrbeu.edu.cn, pengliu@hit.edu.cn, baicj@chinatelecom.cn 
}

\makeatletter
\g@addto@macro\@thanks{
    \footnotetext[1]{Corresponding authors.}
    \footnotetext[2]{Project lead.}
}
\makeatother

\usepackage{bibentry}
\begin{document}

\maketitle

\begin{abstract}
Humanoid robots are promising to learn a diverse set of human-like locomotion behaviors, including standing up, walking, running, and jumping. However, existing methods predominantly require training independent policies for each skill, yielding behavior-specific controllers that exhibit limited generalization and brittle performance when deployed on irregular terrains and in diverse situations. To address this challenge, we propose \emph{Adaptive Humanoid Control (AHC)} that adopts a two-stage framework to learn an adaptive humanoid locomotion controller across different skills and terrains. Specifically, we first train several primary 
locomotion policies and perform a multi-behavior distillation process to obtain a basic multi-behavior controller, facilitating adaptive behavior switching based on the environment. Then, we perform reinforced fine-tuning by collecting online feedback in 
performing adaptive behaviors on more diverse terrains, enhancing terrain adaptability for the 
controller. We conduct experiments in both simulation and real-world experiments in Unitree G1 robots. The results show that our method exhibits strong adaptability across various situations and terrains.
\end{abstract}

\begin{links}
    \link{Website}{https://ahc-humanoid.github.io}
\end{links}

\section{Introduction}

Humanoid robots, due to their human-like morphology, are expected to possess various fundamental human-like locomotion abilities, such as walking, running, and standing up after a fall. Previous studies adopt methods such as whole-body control \cite{sentis2006whole}, model predictive control \cite{li2023multi}, and reinforcement learning (RL) \cite{RL-survey, wang2025more} to enhance the locomotion capabilities of humanoid robots. Recently, RL has achieved a remarkable progress for humanoid locomotion \cite{HumanGym1,HumanGym2,xie2025humanoid}, supported by large-scale simulation \cite{isaac,zakka2025mujoco} and advanced policy gradient methods \cite{PPO1,PPO2}. 

\begin{figure}[t]
\centering
\includegraphics[width=1.0\linewidth]{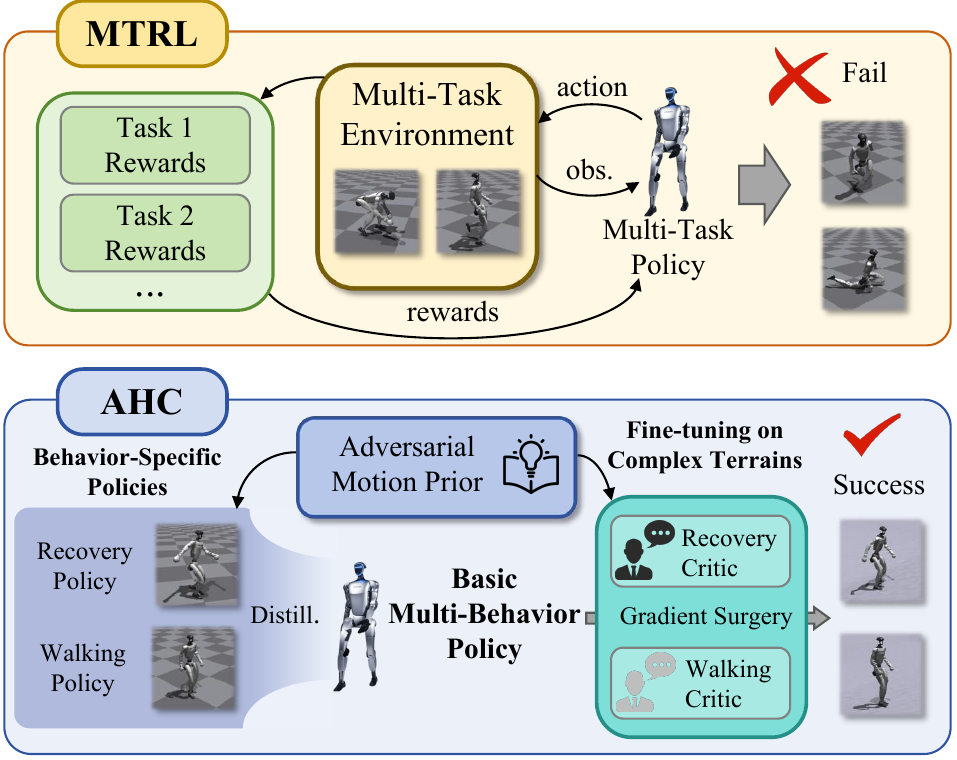}
\caption{Comparison between multi-task RL and our proposed framework. Directly learning multiple skills via multi-task RL is challenging. Therefore, we adopt a two-stage framework consisting of behavior distillation and reinforced fine-tuning, enabling the acquisition of diverse humanoid robot skills and generalization to complex terrains.}
\vspace{-0.2cm}
\label{fig:overview}
\end{figure}

\begin{figure*}[t!]
    \centering
    \includegraphics[width=1.0\linewidth]{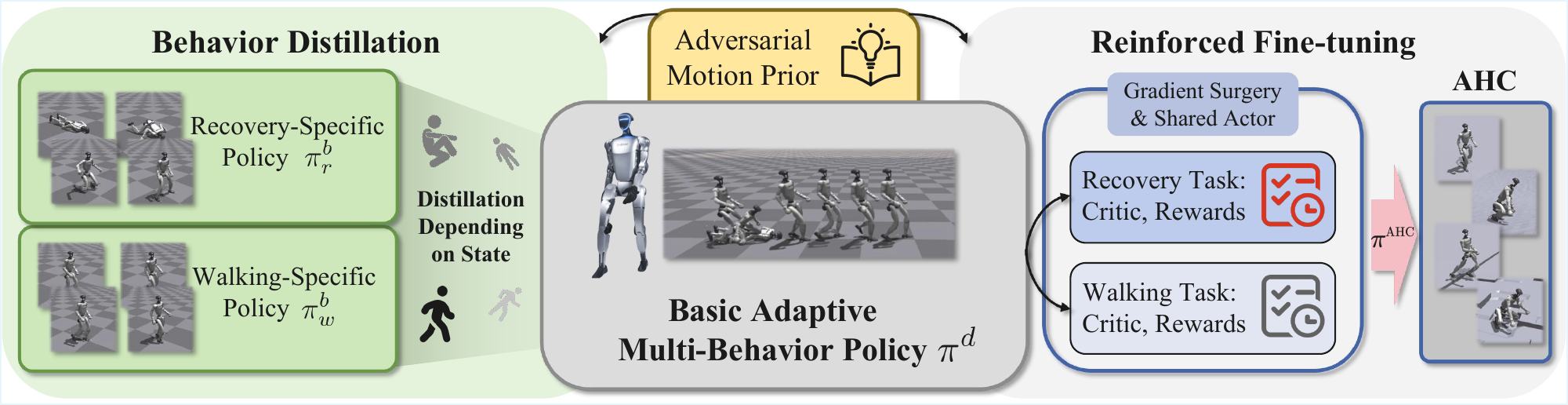}
    \caption{Overview of the proposed two-stage framework \emph{Adaptive Humanoid Control}. In the first stage, we train two separate primary policies on flat terrain. These policies are then distilled into a basic multi-behavior policy via distillation. In the second stage, we perform reinforced fine-tuning on the distilled policy, employing gradient surgery to alleviate gradient conflicts and utilizing behavior-specific critics to provide more accurate value estimation.}
    \label{fig:method}
\end{figure*}

However, existing locomotion algorithms predominantly require training an independent policy with a separately designed reward function, such as standing up \cite{standup1,standup2}, jumping \cite{jump}, squatting \cite{homie}, and walking \cite{HumanGym1}. Such a training paradigm yields independent controllers that excel in performing specific skills, while they exhibit limited generalization abilities in both behavior diversity and terrain adaptability. It is noted that directly extending the learning paradigm to multi-skill setting is challenging, which primarily arises from the conflicts of policy gradients caused by different reward functions \cite{yu2020gradient,liu2021conflict}, hindering the convergence of multi-skill policies. 

Although several works focus on quadruped robot multi-skill learning using Mixture-of-Experts (MoE) or policy distillation \cite{wang2025skill, Yang_2020, huang2025moeloco}, multi-skill learning for humanoid robots still remains challenging and requires further exploration \cite{Parkour, long2024}.

In this work, we propose a novel \emph{Adaptive Humanoid Control (AHC)} method, as illustrated in Figure~\ref{fig:overview}, which adopts a two-stage training framework that first learns a basic multi-behavior controller and then enhances its terrain adaptability to achieve adaptive humanoid control. 
In the first stage, we integrate human motion priors via Adversarial Motion Prior (AMP) with independent behavior-specific policy learning resulting several human-like basic controllers.
Then, we adopt a policy distillation process to obtain a basic multi-behavior controller, which facilitates adaptive behavior switching based on the robot's state and also addresses the challenges in direct RL training of the multi-behavior policy. 
In the second stage, the basic controller collects trajectories and reward feedback in performing different behaviors in different terrains, which is used for continuously improving the terrain adaptability of each behavior via a RL tuning process. We divide the skills into separate tasks and formulate the training as a multi-task RL problem and adopt gradient projection to mitigate the impact of gradient conflicts.

We evaluate the obtained terrain-adapted multi-behavior policy in both simulation and a real-world Unitree G1 robot. Experimental results show that our controller adapts effectively to changes in environmental state (e.g. can stand up after falling and walk) and can perform robust locomotion on challenging terrains (e.g., stairs and slope). 

Our contributions are summarized as follows:
(i) Instead of direct training multi-behavior RL policy across diverse terrains, we employ motion-guided policy learning and supervised distillation to obtain the basic multi-behavior policy. (ii) We adopt sample-efficient RL fine-tune on the basic policy to obtain terrain-adaptive behaviors. (iii) We conduct extensive experiments to verify the learned controller is applicable in both simulation and real-world experiments. 

\section{Methodology}
\subsection{Preliminaries and Problem Definitions}
In the first stage of \emph{AHC}, each behavior-specific humanoid control problem is defined as a Markov Decision Process (MDP) defined as $\mathcal{M} = (\mathcal{S}, \mathcal{A}, \mathcal{P}, R, \gamma)$, where $\mathcal{S}$ and $\mathcal{A}$ are the state space and action space respectively, $\mathcal{P}(\cdot|\bm s, \bm a)$ is the state transition function, $R: \mathcal{S} \times \mathcal{A} \rightarrow \mathcal{R}$ is the reward function and $\gamma \in [0, 1)$ is the reward discount factor. During training, the behavior-specific policy $\pi^{b}$ learns to take a state $\bm s_t$ from $\mathcal{S}$ to output an action $\bm a_t$ to maximize the discounted return $\mathbb{E}\left[\sum_{t=1}^T\gamma^{t-1}R(\bm s_t, \bm a_t)\right]$. The adaptive humanoid control problem is formulated as a multi-task RL problem where each task can be seen as a MDP $\mathcal{M}_{\text{i}} = (\mathcal{S}_{\text{i}}, \mathcal{A}_{\text{i}}, \mathcal{P}_{\text{i}}, R_{\text{i}}, \gamma_{\text{i}})$, $\text{i}\in[1,N]$. Since all tasks (i.e., behaviors) are conducted under a unified environment setup and the controller needs to perform different behavior according to the states ($\mathcal{S} = \bigcup_{\text{i}} \mathcal{S}_{\text{i}}, \mathcal{S}_{\text{i}} \cap \mathcal{S}_{\text{j}} = \emptyset \text{ for } \text{i} \ne \text{j}$), each MDP differs only in its reward function $R_{\text{i}}$ and state space $\mathcal{S}_{\text{i}}$. The objective that behavior-adaptive policy needs to optimize is 

\begin{equation}
\sum_{i=1}^{N}\mathbb{E}_{\mathcal{P},\pi_{i}}\left[\sum_{t=1}^T\gamma^{t-1}R_{i}(\bm s^{i}_t, \bm a_t)\right], \bm s^{i}_t \in \mathcal{S}_{i}.
\end{equation}

% Our problem setting
Behavior-specific policies take privileged information $s^{\text{priv}}_t$ and robot proprioception $s^{\text{prop}}_t$ as inputs since they are not directly deployed in a real robot. 
In the distillation process, we distill the knowledge of behavior-specific policies ($\pi^{\text{b}}_{\text{1}}, \pi^{\text{b}}_{\text{2}}$, ...) into a basic multi-behavior policy $\pi^{\text{d}}$, which aims to perform adaptive behaviors using only proprioception $s^{\text{prop}}_t$ according to different situations (e.g. recovering from fails and walking).
The distilled policy for RL fine-tuning using only $s^{\text{prop}}_t$ to predict the action $\bm a_t$ for robot control by a PD controller. 

The robot proprioception $s^{\text{prop}}_t$ consists of the following parts: 
\begin{equation}
s_t^{\text{prop}} = [\bm{\omega}_t, \bm g_t, \bm c_t, \bm q_t, \bm{\dot{q}}_t, \bm a_{t-1}] \in \mathbb{R}^{69},
\end{equation}
where $\bm{\omega}_t \in \mathbb{R}^3$ is the base angular velocity, $\bm g_t \in \mathbb{R}^3$ is the gravity vector in the base frame, $\bm c_t \in \mathbb{R}^3$ is the velocity command comprising linear velocities along the $x$- and $y$-axes and an angular velocity around the $z$-axis, $\bm q_t \in \mathbb{R}^{20}$ and $\bm{\dot{q}_t} \in \mathbb{R}^{20}$ represent joint position and joint velocity, and $\bm a_{t-1} \in \mathbb{R}^{20}$ is the last action. The privileged information $s^{\text{priv}}_t$ includes ground friction coefficient, motor controller gains, base mass and center of mass shift. The action $\bm a_t$ is converted into a PD target by $\bm q^{\text{target}}_t = \bm q^{\text{default}} + \alpha \bm a_t$ which is used to compute the motor torques:

\begin{equation}
\bm \tau_t = K_p \cdot (\bm q^{\text{target}}_t - \bm q^{\text{default}}) - K_d \cdot \bm{\dot{q}_t}, 
\end{equation}
where $K_p$ and $K_d$ are the stiffness and damping coefficients of the PD controller, and $\alpha$ is a scalar to bound the action.

% The remainder of this section, what key problem this paper solves
The remainder of this section is organized as follows: First, we introduce two fundamental behavior-specific policies: one for robust recovery from various fallen postures and another for human-like locomotion on flat terrain. These policies are then utilized for distilling a basic multi-behavior policy. Second, to enhance terrain adaptability, the distilled multi-behavior policy is fine-tuned across diverse terrains, enabling adaptive learning of both walking and recovery behaviors on more complex terrains. An overview of this two-stage framework is illustrated in Figure~\ref{fig:method}.

\subsection{Multi-Behavior Distillation}
Training an adaptive multi-behavior controller from scratch via online RL is challenging, as diverse behaviors require distinct environment settings and reward functions, often leading to poor policy convergence \cite{sodhani2021multitaskrl, liu2021conflict}.
This is primarily caused by issues such as gradient conflict and gradient imbalance. To address this issue and enable the policy to learn multiple primary behaviors, we first train two basic behavior-specific policies using Proximal Policy Optimization (PPO) \cite{PPO1}, which are then distilled into a basic multi-behavior policy $\pi^d$.

\paragraph{Falling Recovery Behavior Policy $\pi^{b}_{r}$}
Recovering from unexpected falls is an essential capability for ensuring the robustness and autonomy of humanoid robots. HoST \cite{standup2} has shown that a standing-up control policy can be trained via multiple critics and force curriculum. We also employ multiple critics \cite{mysore2022multi} for our basic standing-up policy $\pi^{b}_{r}$. In the surrogate loss for policy gradient, the advantage function is estimated using a weighted formulation $\hat{A} = \sum_{i=0}^{n}\omega_i \cdot (\hat{A}_i - \mu_{\hat{A}_i})/\sigma_{\hat{A}_i}$, where $\omega_i$ is a weighting coefficient, and $\mu_{\hat{A}_i}$ and $\sigma_{\hat{A}_i}$ correspond to the batch mean and standard deviation of the advantage from group $i$, respectively. 

Different from HoST, we let the policy focus on recovery on flat terrains. Specifically, the robot is initialized in either a supine or prone position, with additional joint initialization noise introduced to encourage learning robust recovery behaviors from various postures. To mitigate the negative impact caused by interference between sampled rollouts from different initial postures \cite{standup2} and to promote more natural standing-up motions, we introduce AMP-based \cite{AMP, escontrela2022adversarial} reward function using a discriminator which determines whether an episode is a positive sample (i.e., from the reference motion) or a negative sample (i.e., from the robot). The output of the discriminator can be used to guide the robot to recover in a smooth and plausible manner. Consequently, the standing-up policy $\pi^{b}_{r}$ can robustly recover from diverse abnormal postures and learns behaviors from the reference motion, such as leveraging the arms to support the ground during the standing-up process.
The AMP-based reward formulation and the objective of the discriminator can be found in Appendix A.

\paragraph{Walking Behavior Policy $\pi^{b}_w$}
Locomotion across diverse terrains is also essential for humanoid robots. Although complex system designs and multi-stage training pipelines can enable robots to learn diverse locomotion skills \cite{hwcloco, Parkour}, we only adopt a simple framework and reward functions design to enable the policy $\pi^{b}_{w}$ to learn fundamental locomotion ability. 

$\pi^{b}_{w}$ is capable of walking on flat terrain in response to a velocity command $\bm c_t$.
To enable the robot to move in a human-like manner and accelerate the convergence process, an AMP-based reward function is introduced similar to that of $\pi^{b}_{r}$. It is worth noticing that although the policy $\pi^{b}_{w}$ is only capable of walking, after undergoing distillation and RL fine-tuning, our policy is able to adapt to diverse terrains while exhibiting significantly improved robustness to external disturbances. The PPO parameters, reward designs, network architecture and domain randomization terms for the basic behavior policies training are listed in Appendix B.

\paragraph{Behavior Distillation}
In the behavior distillation process, we use DAgger \cite{chen2020learning, ross2011reduction} to distill different behavior knowledge into an MoE-based multi-behavior policy $\pi^{d}$, which eliminates the gradient conflicts resulting from different reward landscapes of various behaviors. 

The MoE module can automatically assign different experts to learn distinct behaviors, enabling the policy to leverage such prior knowledge for efficient multi-behavior improvement and multi-terrain adaptability in subsequent RL fine-tuning stage.
During the training process, the robot is initialized in a fallen or standing posture and the policy is supervised by $\pi^b_r$ or $\pi^b_w$ according to which behavior it should perform. The loss function of $\pi^d$ is computed by: 
\begin{equation}
\mathcal{L}_{\pi^d}(\bm s_t) = \left\{
\begin{array}{lcl}
\mathbb{E}_{s_t, \pi_d, \pi_r^b}\left[\|\bm a^{\pi^{d}}_t-\bm a^{\pi^b_r}_t\|_2^2\right],       &      & \bm s_t \in \mathcal{S}_r\\
\mathbb{E}_{s_t, \pi_d, \pi_w^b}\left[\|\bm a^{\pi^{d}}_t-\bm a^{\pi^b_w}_t\|_2^2\right],     &      & \bm s_t \in \mathcal{S}_w
\end{array} \right.
,
\end{equation}
where $\bm a^{\pi^d}_t$, $\bm a^{\pi^b_r}_t$ and $\bm a^{\pi^b_w}_t$ are sampled from their corresponding policy, $\mathcal{S}_r$ and $\mathcal{S}_w$ represent standing-up and walking state space. The distillation process employs the same domain randomization as the behavior-specific policy training and $\pi^d$ takes only proprioception as input. The distillation process not only integrates the basic behaviors into a single policy but also enhances each of them individually. Specifically, $\pi^d$ exhibits more robust walking performance, as it learns to recover from near-fall postures, and demonstrates a more natural standing pose after a stand-up behavior, which facilitates smoother transitions into walking.

\subsection{RL Fine-Tuning}
In the RL fine-tuning stage, we formulate the problem as a multi-task RL problem, where the policy $\pi^{ft}$ is initialized with the distilled policy $\pi^d$ from last stage and learns fail recovery task and walking task on complex terrains. 
To maintain human-likeness, we also adopt an AMP-based reward in this stage using the same reference motion as in the previous stage. Leveraging the MoE module and prior knowledge in basic multi-behavior policy $\pi^d$, the gradient conflict problem is alleviated, enabling efficient learning of adaptive behaviors on various terrains.
We fine-tune the policy $\pi^{ft}$ using PPO on two GPUs, where each GPU handles one task under its corresponding environment setup. The policies for different task share the same set of parameters.

\paragraph{Behavior-Specific Critics and Shared Actor}
In the PPO algorithm, the policy gradient is computed using normalized advantages, which stabilizes the surrogate loss by standardizing the scale of the advantage estimates. However, the value loss relies on unnormalized return targets, since the critic must approximate the true expected return to provide meaningful value estimates.

To prevent tasks with larger reward scales from dominating gradient updates and hindering others’ learning \cite{chen2018gradnorm, hessel2019multi}, we use behavior-specific critics with a shared actor during fine-tuning.
By assigning a separate critic to each task, we isolate value function learning for task-specific reward functions which enables more accurate value estimation and allows for customized critic architectures for each task (e.g., multiple critics for standing-up behavior).
Meanwhile, a shared actor is updated using policy gradients aggregated across tasks, allowing skill transfer and terrain adaptability.

\paragraph{Eliminating Gradient Conflict in Multi-task Learning}
While the use of behavior-specific critics addresses discrepancies in gradient magnitude, the shared actor still aggregates potentially conflicting gradients from different tasks.

To alleviate this issue, we apply Projecting Conflicting Gradients (PCGrad) \cite{yu2020gradient} to resolve gradient conflicts during optimization. For each pair of task gradients, if they conflict (i.e., their cosine similarity is negative), the gradient of one task is projected onto the normal plane of the other, removing the conflicting direction while preserving progress on the remaining subspace. Specifically, given two task gradients $\mathbf{g}_i$ and $\mathbf{g}_j$, the projected gradient is computed as:
\begin{equation}
    \mathbf{g}_i=\mathbf{g}_i-\frac{\mathbf{g}_i\cdot \mathbf{g}_j}{\left \| \mathbf{g}_j \right \|^2 }\mathbf{g}_j 
\end{equation}
In practice, we integrate PCGrad before the actor update step: each task computes its local actor gradient on its dedicated GPU, and all gradients are then communicated to the main process, where PCGrad is applied to perform gradient surgery. 
After performing the optimizer step with the conflict-free gradients on the main process, we broadcast the updated parameters back to all workers. 
PCGrad allows the shared actor to learn multiple task-specific skills without gradient conflicts, ensuring efficient multi-task RL learning. The detailed process can be found in Appendix B.

\paragraph{Terrain Curriculum}
Following the previous work \cite{rudin2022learning}, we adopt the terrain curriculum to improve learning efficiency and adaptability on diverse terrains.
An automatic level adjustment mechanism modulates terrain difficulty based on task-specific performance. We design flat, slope, hurdle and discrete terrains for both tasks. The maximum slope inclination is 16.6°. Hurdle terrains are composed of regularly spaced vertical obstacles with a maximum height 0.1$m$. The discrete terrain consists of randomly placed rectangular blocks. Specifically, we generate 20 rectangular obstacles with width and lengths sampled between 0.5$m$ and 2.0$m$, and heights uniformly sampled between 0.03$m$ and 0.15$m$. We arrange the terrain map into a 10 × 12 grid of 8$m\times$ 8$m$ patches, with 10 difficulty levels and 3 columns per terrain type.

\section{Experimental Results and Discussion}
We perform training in the IsaacGym simulator, using 4096 parallel environments. We train the behavior-specific policies for 10,000 iterations, followed by 4,000 iterations of policy distillation. Fine-tuning for terrain adaptability runs an additional 10,000 iterations using online RL on two NVIDIA RTX 4090 GPUs. The 20-DoF action space (excluding waist joints) is used, with AMP rewards from retargeted motion capture for recovery and LAFAN1 data for locomotion. The policy is deployed on a Unitree G1 humanoid robot at 50Hz, tracked by a 500Hz PD controller converting joint positions to torques.
\subsection{Simulation Results}
\paragraph{Terrain Adaptability}
\begin{table*}[t!]
\centering
\caption{Performance evaluation on locomotion and fail recovery tasks across different terrains. To compare our approach with the baselines, we report the traversal success rate (Succ.) and average traversing distance (Dist.) for the locomotion task, and the success rate (Succ.) for the fail recovery task.}
\resizebox{2\columnwidth}{!}{
\begin{tabular}{c|cc|cc|cc|cc|c|c|c|c@{}}
\toprule
\multirow{3}{*}{\textbf{Method}} & 
\multicolumn{8}{c|}{\textbf{Locomotion}} & 
\multicolumn{4}{c}{\textbf{Fail Recovery}} \\
\cmidrule(l){2-9} \cmidrule{10-13}
 & \multicolumn{2}{c|}{Plane} & \multicolumn{2}{c|}{Slope} & \multicolumn{2}{c|}{Hurdle} & \multicolumn{2}{c|}{Discrete} & Plane & Slope & Hurdle & Discrete \\
 & Succ. & Dist. & Succ. & Dist. & Succ. & Dist. & Succ. & Dist. & Succ. & Succ. & Succ. & Succ. \\
\midrule
HOMIE \cite{homie}        & 0.802 & 6.421 & 0.599 & 4.795 & 0.407 & 3.259 & 0.442 & 3.603 & - & - & - & - \\
HoST \cite{standup2}         & \multicolumn{2}{c|}{-} & \multicolumn{2}{c|}{-} & \multicolumn{2}{c|}{-} & \multicolumn{2}{c|}{-} & 0.997 & 0.978 & 0.911 & 0.843 \\
\midrule
Fail Recovery Policy        & \multicolumn{2}{c|}{-} & \multicolumn{2}{c|}{-} & \multicolumn{2}{c|}{-} & \multicolumn{2}{c|}{-} & \textbf{1.000} & 0.757 & \textbf{0.999} &  0.942 \\
Walking Policy         & 0.971 & 7.782 & 0.000 & 2.197 & 0.161 & 2.561 & 0.127 & 3.788 & - & - & - & - \\
Multi-Behaviors Policy         & \textbf{0.993} & \textbf{7.969} & 0.160 & 5.850 & 0.756 & 7.254 & 0.702 & 7.321 & 0.999 & 0.904 & 0.997 & 0.947 \\
AHC          & 0.992 & 7.951 & \textbf{0.975} & \textbf{7.987} & \textbf{0.922} & \textbf{7.866} & \textbf{0.969} & \textbf{7.951} & \textbf{1.000} & \textbf{1.000} & 0.985 & \textbf{0.969} \\
\bottomrule
\end{tabular}}
\label{table:performance}
\end{table*}

To evaluate the terrain-adaptive capability of our proposed \emph{AHC} framework, we compare it with the following methods: (1) \textbf{HOMIE} \cite{homie}: We adapt the lower-body locomotion policy from HOMIE to our setting by re-training it on our terrain settings. (2) \textbf{HoST} \cite{standup2}: We train the standing-up controller from HoST on our terrain settings. (3) \textbf{Fail Recovery Policy $\pi_r^b$}: The fail recovery behavior policy trained in the first stage of our framework. (4) \textbf{Walking Policy $\pi_w^b$}: The walking behavior policy trained in the first stage of our framework. (5) \textbf{Basic Multi-Behaviors Policy $\pi^d$}: The basic multi-behavior policy from the first stage distillation process.

We construct four types of terrain patches for evaluation, each of size 8 $m\times$ 8 $m$, corresponding to the four terrain types used during training: flat, slope, hurdle, and discrete. On the slope terrain, the inclination angle is uniformly sampled between 12° and 16°. The hurdle obstacle heights uniformly sampled between 0.08$m$ to 0.1$m$. For the locomotion task, the hurdle terrain includes 3 obstacles, while the recovery task uses a more cluttered setup with 8 obstacles. The discrete terrain consists of randomly positioned rectangular blocks, with height variations ranging from 0.08$m$ to 0.1$m$. Each policy is evaluated separately on these terrain types to assess its performance.

We adopt \textbf{Success Rate (Succ.)} as the primary metric for both tasks. For the locomotion task, success rate is defined as the percentage of trials in which the robot traverses the full 8$m$ terrain within 20$s$ without termination. During evaluation, the robot is assigned a fixed forward velocity at the beginning of each episode, uniformly sampled from 0.4$m/s$-1.0$m/s$. An episode is terminated if the robot walks off the current 8$m \times $8$m$ terrain patch or falls irrecoverably. For policies that integrate both fail recovery and locomotion capabilities (i.e., multi-behavior policy $\pi^d$ and \emph{AHC} policy $\pi^{\text{AHC}}$), falling does not trigger termination, allowing the robot to autonomously recover and resume traversal. For recovery task, a trial is considered successful if the robot can stand up from a fallen posture and maintain balance without falling again within 10$s$. In addition, we report \textbf{Traversing Distance (Dist.)}, defined as the average distance covered before termination, computed over all trials, including both successful and failed ones. All evaluations are conducted using 1000 parallel environments.

As shown in Table~\ref{table:performance}, our proposed \emph{AHC} policy $\pi^{\text{AHC}}$ outperforms both HOMIE and HoST on the locomotion and recovery tasks across a wide range of terrain types. In locomotion task, the performance gain is largely attributed to $\pi^{\text{AHC}}$'s ability to autonomously recover from falls and resume traversal, especially on terrains with high obstacle density (e.g., hurdle and discrete terrains). In addition, the incorporation of AMP provides motion priors that guide the policy toward learning stable and robust behaviors, contributing to $\pi^{\text{AHC}}$'s better performance compared to HoST. We further compare the multi-behavior policy $\pi^d$ with the behavior-specific policies (i.e., $\pi^b_r$ and $\pi^b_w$). The multi-behavior policy $\pi^d$ demonstrates superior robustness in the locomotion task on complex terrains such as hurdle and discrete. This improvement stems from its seamless integration of walking and recovery behaviors within a single policy. These results underscore the promise of integrating complementary skills such as walking and recovery into a unified policy to achieve robust locomotion in challenging environments. To further enhance terrain adaptability, we apply RL to fine-tune $\pi^d$ on diverse terrains, resulting in the final \emph{AHC} policy $\pi^\text{AHC}$. This additional stage brings improvements on most terrain types across both tasks, particularly in slope and discrete terrains, highlighting the transferability of our two-stage training framework.

\begin{figure}[t]
    \centering
    \includegraphics[width=1.0\linewidth]{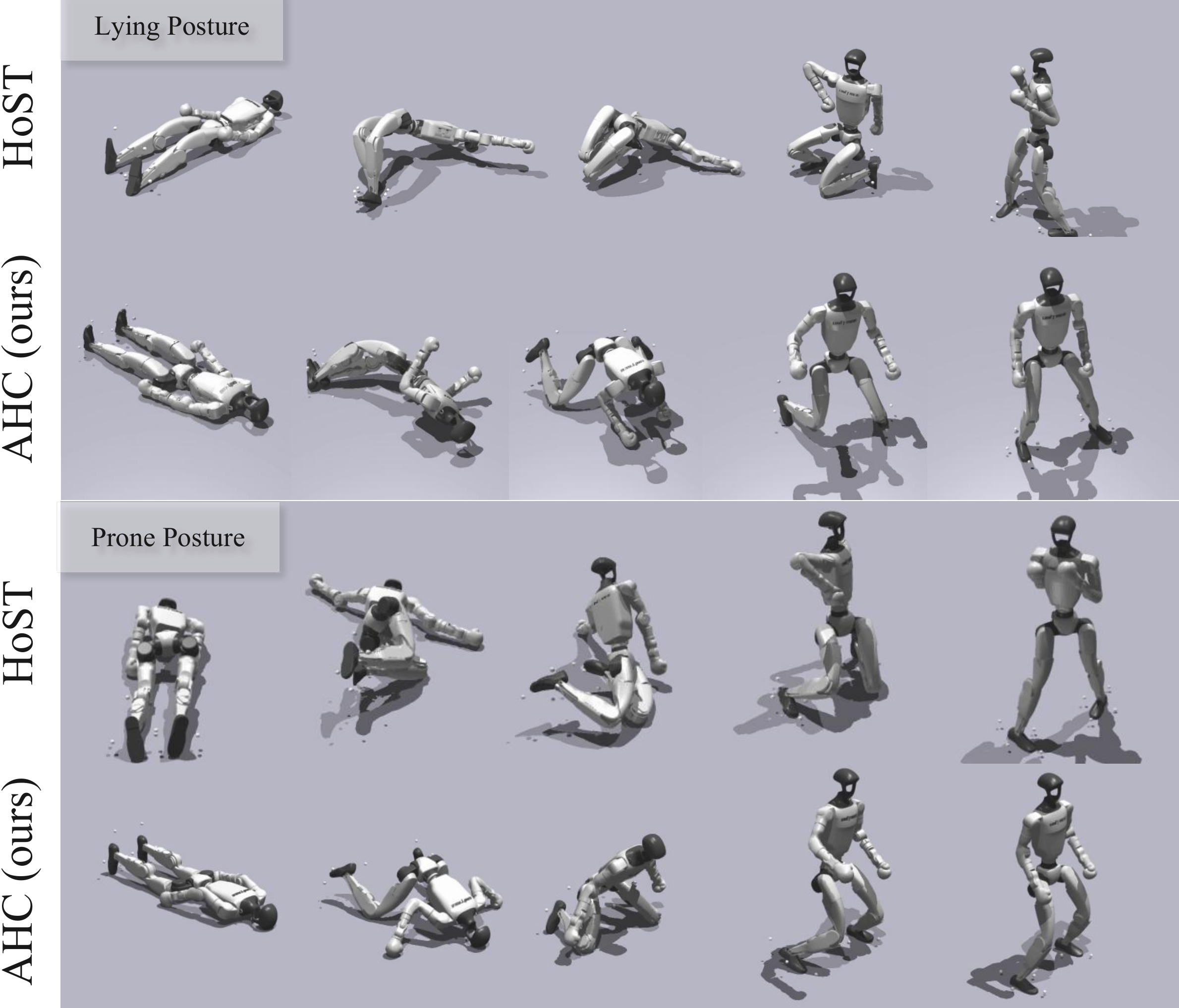}
    \caption{Comparison of recovery motions under AHC and HoST. We compare our AHC (with AMP) against the HoST (w/o AMP) in both lying and prone scenarios. AHC produces smoother recovery behaviors. This highlights the effectiveness of AMP in guiding the learning of naturalistic recovery motions.}
    \label{fig:amp_exp}
\end{figure}

\begin{figure}[t]
    \centering
    \includegraphics[width=1.0\linewidth]{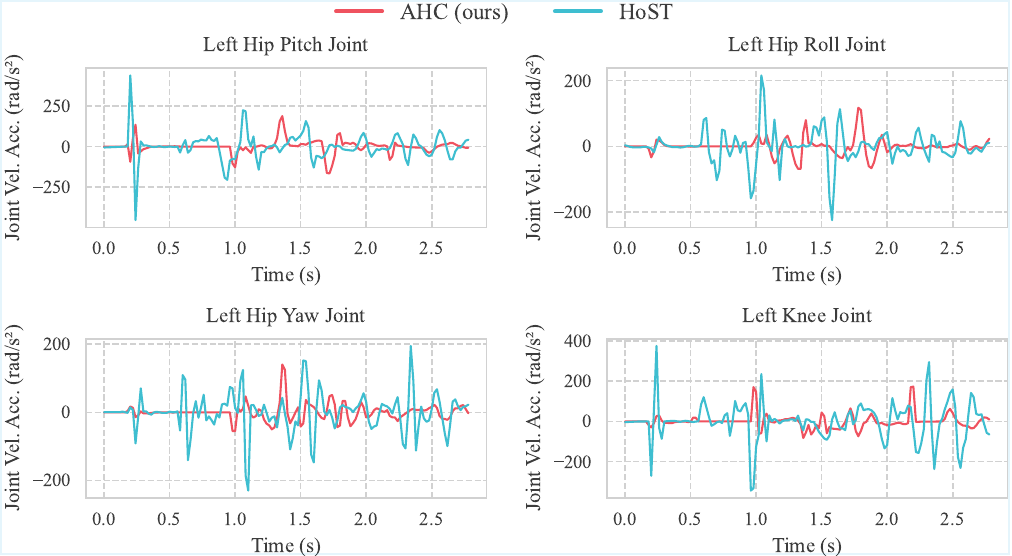}
    \caption{Joint acceleration analysis of the left leg during recovery. Acceleration profiles of hip and knee joints from the left leg illustrate that our AHC results in stable joint actuation, with notably fewer abrupt fluctuations compared to HoST.}
    \vspace{-0.2cm}
    \label{fig:amp_joint_acc}
\end{figure}

\paragraph{AMP for Standing-Up}
We compare \emph{AHC}, which incorporates AMP, with the HoST \cite{standup2} baseline without human motion priors. As shown in Figure~\ref{fig:amp_exp}, we visualize a sequence of snapshots during the stand-up process under both approaches in lying and prone posture. Without AMP (i.e., HoST), the robot exhibits uncoordinated and jerky motions, relying on abrupt limb movements to return to a standing posture. In contrast, \emph{AHC} policy generates a natural get-up motion, including leg folding, arm support, and trunk lifting. We further compare the joint velocity accelerate during the recovery. As shown in Figure~\ref{fig:amp_joint_acc}, the velocity curves of key joints (i.e., hip and knee) for \emph{AHC} policy exhibit fewer abrupt fluctuations compared to the policy trained with HoST. These results demonstrate that AMP helps shape the recovery controller towards producing stable motions, which are difficult to obtain with handcrafted reward functions.

\begin{table}[h]
\centering
\caption{Gradient Cosine Similarity between Tasks across Different Ablation Settings. A lower similarity indicates higher conflict between task gradients.}
\begin{tabular}{l|c}
\toprule
\textbf{Method} & \textbf{Cosine Similarity ($\uparrow$)} \\
\midrule
AHC-SC-w/o-PC & 0.247 \\
AHC-SC-PC    & \textbf{0.519} \\
\midrule
AHC-BC-w/o-PC & 0.334 \\
AHC (ours) & \textbf{0.535} \\
\bottomrule
\end{tabular}
\label{table:grad_cos_sim}
\end{table}

\paragraph{Ablation on PCGrad and Behavior-Specific Critics}
We conduct a comprehensive ablation study to evaluate the contributions of the two key components introduced in the second-stage fine-tuning stage: PCGrad and behavior-specific critics update strategy. We examine four configurations: (1) \textbf{AHC-SC-w/o-PC}: a single shared critic without PCGrad. (2) \textbf{AHC-SC-PC}: a single shared critic with PCGrad. (3) \textbf{AHC-BC-w/o-PC}: behavior-specific critics without PCGrad. (4) \textbf{AHC (Ours)}: behavior-specific critics with PCGrad. In the single critic setting, a shared critic network is jointly optimized across all tasks.

\begin{figure}[t]
    \centering
    \includegraphics[width=1.0\linewidth]{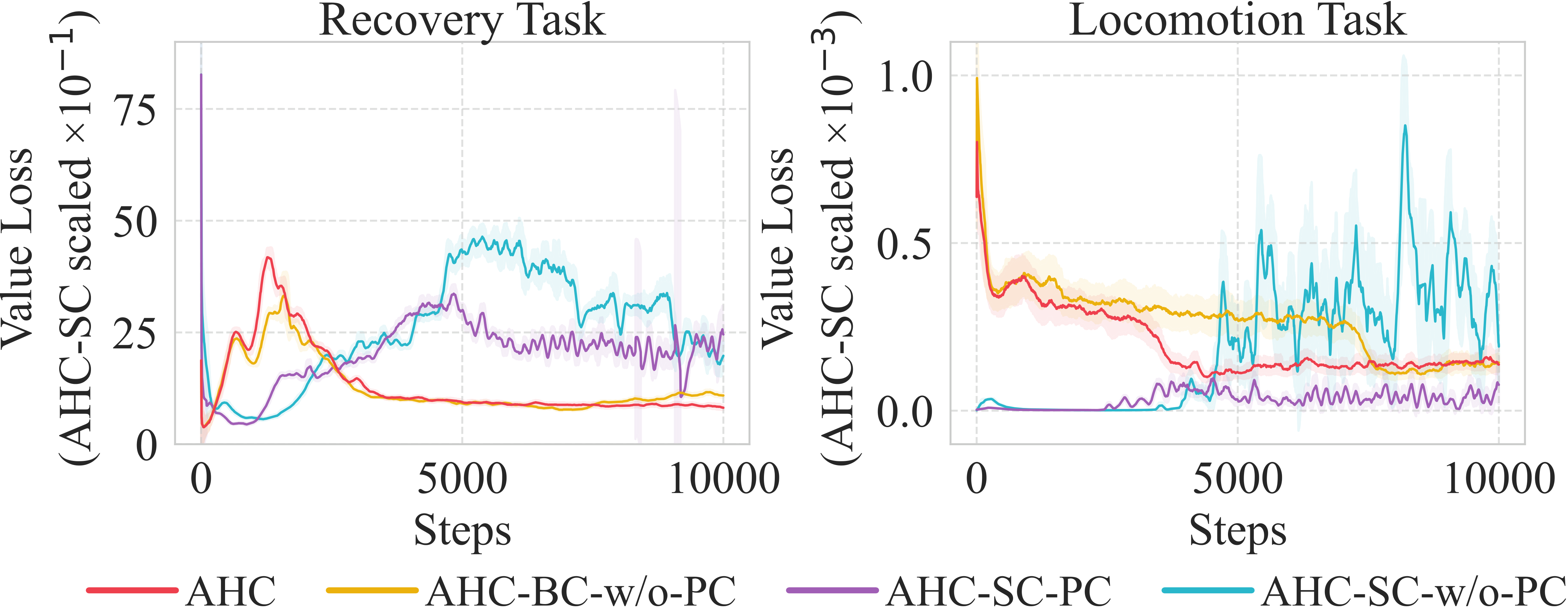}
    \caption{Value loss curves during the second-stage fine-tuning. Policies equipped with behavior-specific critics (AHC-BC-w/o-PC and \emph{AHC}) indicate more stable value learning compared to their shared-critic counterparts (AHC-SC).}
    \label{fig:value_loss}
\end{figure}

\begin{figure}[t]
    \centering
    \includegraphics[width=1.0\linewidth]{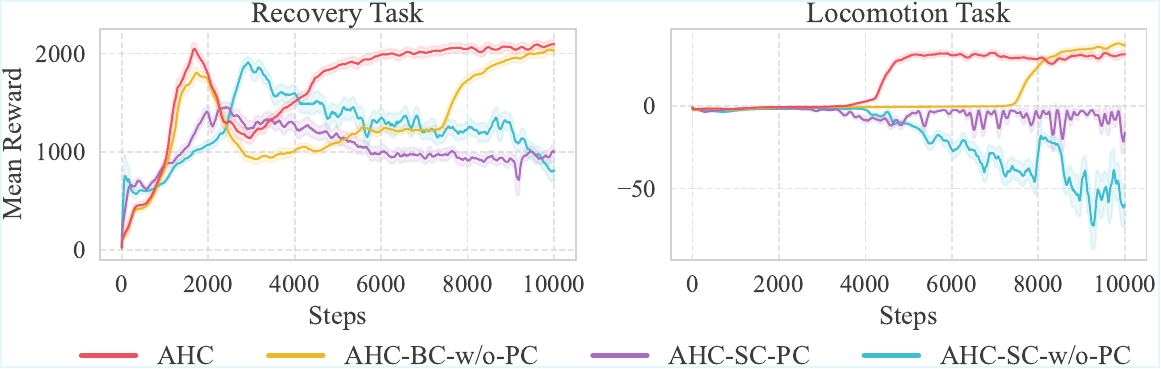}
    \caption{Training episode return curves during second-stage fine-tuning. With PCGrad and behavior-specific critics \emph{AHC} achieve higher and more balanced returns across tasks.}
    \label{fig:episode_return}
\end{figure}

To quantify the role of PCGrad in mitigating gradient conflicts during multi-task optimization, we compute the average cosine similarity between the gradients of the two tasks during the second-stage training. As shown in Table~\ref{table:grad_cos_sim}, PCGrad reduce gradient conflict, resulting in higher cosine similarity values in both critic settings. Notably, the use of behavior-specific critics also yield higher similarity, indicating that they help alleviate gradient conflicts between tasks. We further investigate the impact of adopting behavior-specific critics by monitoring the evolution of value loss during training. As illustrated in Figure~\ref{fig:value_loss}, models with behavior-specific critics (AHC-BC-w/o-PC and AHC) achieve lower value loss compared to their shared-critic counterparts. This suggests that decoupling value learning for each task helps mitigate optimization difficulties caused by reward scale discrepancies. In addition, we visualize the training episode return curves in Figure~\ref{fig:episode_return} to evaluate how well each configuration balances learning across tasks. The shared-critic variants (AHC-SC) tend to neglect the locomotion task due to its smaller reward magnitude. In contrast, \emph{AHC} maintains high returns for both tasks, demonstrating superior performance in multi-task learning. Notably, AHC exhibits faster convergence compared to the other settings. These results highlight the effectiveness of incorporating both PCGrad and behavior-specific critics in facilitating balanced and efficient optimization during the second stage.

\begin{figure*}[t]
    \centering
    \includegraphics[width=1.0\linewidth]{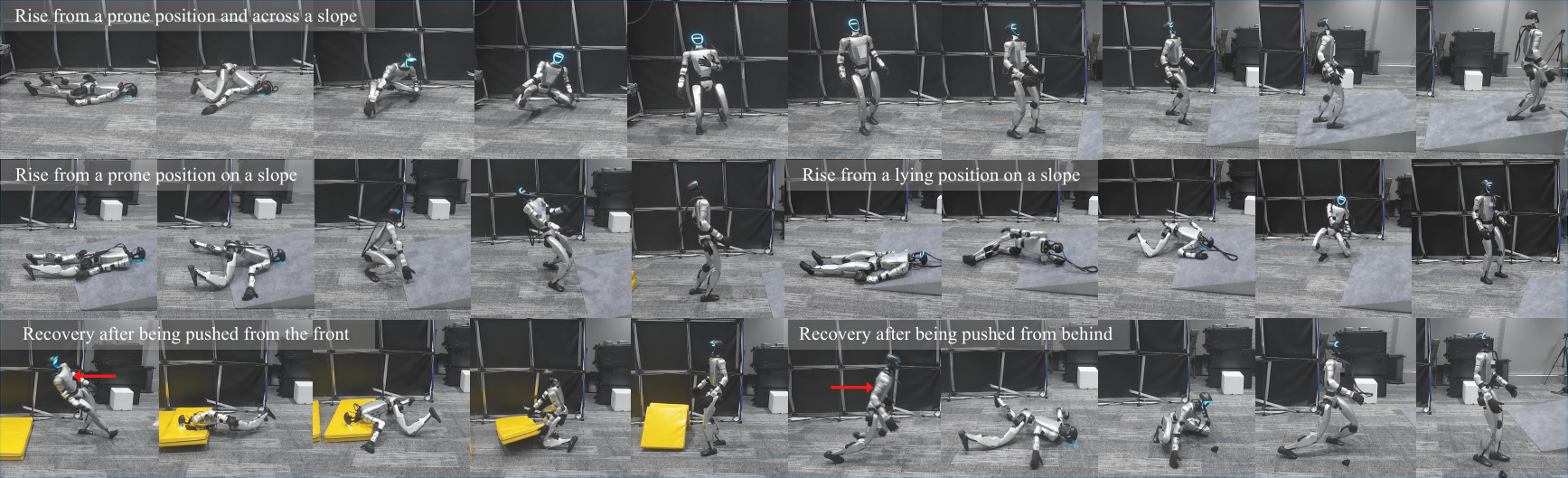}
    \caption{Snapshot of real-world deployment. The robot performs recovery and locomotion in diverse scenarios, including standing up from prone and lying positions on sloped terrain and recovering after external pushes during walking.}
    \label{fig:deploy}
\end{figure*}

\subsection{Deployment Results}
We deploy our trained policy to a Unitree G1 humanoid robot in real-world settings without additional fine-tuning, validating its effectiveness across diverse scenarios. A sequence of deployment snapshots is shown in Figure~\ref{fig:deploy}. To evaluate robustness and generalization, we conduct multiple trials in our laboratory environment under different conditions. For the recovery evaluation, we place the robot in both supine and prone init position on flat ground and inclined terrain. In all cases, the robot successfully recovers from various fallen postures, including moderate external disturbances. After each recovery, it stabilizes itself and smoothly transitions into a walking-ready posture, displaying natural and coordinated motion. For the locomotion task, we test the policy in two initialization settings: recovery followed by walking, and directly starting from a standing posture. In both cases, the robot is able to walk stably on flat ground and inclined surfaces, demonstrating robust control and effective tracking of velocity commands. During walking, we apply external pushed in random directions to assess the robot's ability to maintain balance. The robot generally withstands perturbations and continues walking. Even when a fall does occur during walking, the robot is able to autonomously perform the recovery maneuver and resume locomotion across the terrain, exhibiting strong resilience and long-horizon autonomy. These results suggest that the learned policy not only bridges the sim-to-real gap effectively, but also integrate recovery and locomotion behaviors in a cohesive and robust manner.

\section{Related Work}
\paragraph{Humanoid Locomotion}
Deep reinforcement learning (RL) algorithms have enabled humanoid robots to perform robust and even highly complex locomotion behaviors in simulation environments \cite{isaac, todorov2012mujoco}. Without relying on external sensors, prior works have demonstrated several challenging behaviors, including coordinated control of the upper and lower body \cite{almi}, whole-body locomotion \cite{2024realrlloco}, robust traversal over complex terrains \cite{gu2024advancinghumanoidlocomotionmastering}, and highly flexible full-body teleoperation \cite{homie}. Fundamental capabilities such as fall recovery have also been achieved through techniques like using multiple critics \cite{standup2}.
The RL policy with external sensors like depth cameras and LiDARs is able to perceive the environment, like terrains and obstacles, and learn to avoid obstacles and leap a gap \cite{pim, vb-com}. More extreme terrains with sparse footholds can be traversed by employing finer terrain perception or attention-based network designs \cite{wang2025beamdojo, he2025attentionbasedmapencodinglearning}.
However, all these works focus solely on a single behavior, such as moving or recovery. In contrast, our proposed framework enables the robot to acquire multiple skills and autonomously select appropriate behaviors based on its current state.

\paragraph{Multi-Behavior Learning in Robots}
Mastering multiple behaviors is essential for enhancing the adaptability and practical application of robots. Policy distillation allows the integration of skills from multiple expert policies into a single policy, enabling diverse behaviors for navigating complex terrains \cite{robotparkour, Parkour}. Alternatively, hierarchical frameworks can select among multiple skill policies to facilitate efficient multi-skill traversal \cite{Anymalparkour}. HugWBC \cite{xue2025unified} leverages input signals such as gait frequency and foot contact patterns to guide the policy, allowing it to exhibit different behaviors in response to varying commands. MoE-Loco \cite{huang2025moeloco} adopts an MoE architecture to reduce gradient conflicts in multi-skill RL, thereby improving training efficiency. MoRE \cite{wang2025more} further enhances policy performance by incorporating AMP-based rewards and external sensor inputs.
However, prior works typically achieve multi-behavior capabilities either through explicit control signals or by combining behaviors with high similarity (e.g., stair climbing and gap jumping). In contrast, our proposed framework integrates highly diverse behaviors into a single unified policy and enables the robot to autonomously switch between them based on its state.

\section{Conclusion and Future Work}
In this paper, we propose a two-stage framework, \emph{Adaptive Humanoid Control} (\emph{AHC}). The first stage distills a basic multi-behavior policy, while the second stage fine-tunes it for terrain adaptability. The resulting controller enables robust locomotion across diverse terrains and effective recovery from various types of falls. By integrating an MoE architecture, gradient projection techniques, and behavior-specific critics, our approach enhances multi-task learning efficiency and mitigates gradient conflicts. Extensive simulation and real-world experiments validate the robustness and adaptability of the proposed \emph{AHC} policy. Future work will explore augmenting perceptual capabilities with external sensors and expanding the behavior category for even greater generalization.

\section{Acknowledgments}
This work is supported by the National Natural Science Foundation of China (Grant No.62306242, No.62302120),  the Young Elite Scientists Sponsorship Program by CAST (Grant No.2024QNRC001), the Yangfan Project of the Shanghai (Grant No.23YF11462200), and the Heilongjiang Key R\&D Program of China (Grant No.GA23A915).

\bibliography{aaai2026}

\clearpage
\appendix
\section{Appendix}
% DR 
% Reward
% AMP objective & reward & discriminator setup
% Network architecture and parameter
% PPO
% optional

% Terrain 
% different init state setup
% multi-gpu
% Deployment Details Kp Kd

\subsection{A. AMP Reward Formulation and Discriminator Objective}
We adopt Adversarial Motion Prior (AMP)~\cite{AMP,escontrela2022adversarial} to provide a style reward that encourages natural behaviors. The AMP consists of a discriminator $D_{\phi}$ that determines whether a sequence of states originates from the reference motion dataset or from the policy. We construct the AMP input state $s_t^{\text{amp}} \in \mathbb{R}^{20}$ by extracting 20 joint positions from the full observation. Unlike previous works that rely on only two consecutive states, we provide the discriminator with a temporal context by feeding a 5-step window of AMP states. Specifically, the input sequence of discriminator is defined as $\tau_t = (s_{t-3}^{\text{amp}}, s_{t-2}^{\text{amp}}, s_{t-1}^{\text{amp}}, s_t^{\text{amp}}, s_{t+1}^{\text{amp}})$.

Given a reference motion dataset $\mathcal{M}$ and on-policy rollouts $\mathcal{P}$ collected during policy and environment interaction, we construct 5-step motion sequences for training the discriminator. We randomly sample reference sequences using a sliding window over entire motion trajectories in $\mathcal{M}$. The discriminator is trained to assign higher scores to reference sequences and lower scores to policy-generated ones. Its objective is formulated as:
\begin{align}
\label{eq:disc_loss}
\arg\max_{\phi} \ \ & \mathbb{E}_{\tau \sim \mathcal{M}}[(D_{\phi}(\tau) - 1)^2] + \mathbb{E}_{\tau \sim \mathcal{P}}[(D_{\phi}(\tau) + 1)^2] \nonumber \\
& + \frac{\alpha^{d}}{2} \mathbb{E}_{\tau \sim \mathcal{M}}[\|\nabla_{\phi} D_{\phi}(\tau)\|_2],
\end{align}
where the first two terms represent the discriminator loss following the least square GAN formulation, while the third term is a gradient penalty that helps mitigate training instability. $\alpha^d$ is a manually specified coefficient controlling the strength of this regularization. The discriminator output $D_{\phi}(\tau) \in \mathbb{R}$ denotes the scalar score predicted by the discriminator for the state sequence $\tau$. 

Following prior works, we use the discriminator output $d = D_{\phi}(\tau_t)$ to define a smooth surrogate reward function:
\begin{equation}
    r^{\text{style}}(s_t) = \alpha \cdot \max\left(0,\ 1 - \frac{1}{4}(d - 1)^2\right),
\end{equation}
where $\alpha$ is a scaling factor. The total reward used for policy optimization is the sum of the task and style rewards:
\begin{equation}
    r_t = r_t^{\text{task}} + r_t^{\text{style}}.
\end{equation}
This style reward encourages the policy to perform locomotion behavior that closely resemble those in the reference dataset. In our setup, each task (i.e., locomotion and recovery) is associated with an independent discriminator and its corresponding reference motion data.

\subsection{B. Training Details}
% specific: networks, hyper parameters
% in common: dr, kp&kd, 
\label{appendix:training_details}

\paragraph{Multi-Behaviors Distillation Policy}
% behavior-specific policy (teacher), reward functions(loco&recovery), network&hyper-params in common
% distillation policy (student), distillation loss funciton,distillation training hyper-params, network&hyper-params
In the first training stage, we adopt a teacher-student framework. Two behavior-specific policies are independently trained using PPO, each specialized for a single task: recovery or locomotion. These policies serve as teacher policies and are granted access to privileged information available only in simulation, which facilitates more efficient learning. We then distill the learned skills into a basic multi-behavior student policy that performs both skills without relying on privileged inputs.

Each behavior-specific policy adopts a same actor-critic architecture, which includes a history encoder shared between the actor and critic networks. The history encoder process 10 step history observation through a 3-layer MLP with hidden dimensions [1024, 512, 128], outputting a latent embedding of dimension 64. This latent is concatenated with the current observation and passed to both the actor and critic networks. The actor is a 3-layer MLP with hidden sizes [512, 256, 128], outputting mean actions with a learnable diagonal Gaussian standard deviation. The critic consists of $N$ independent networks, each implemented as a 3-layer MLP with hidden dimensions [512, 256], where each critic corresponds to a specific reward group. 

Our PPO implementation follows the standard formulation with clipped surrogate loss and generalized advantage estimation (GAE). We use Adam optimizer with the learning rate of $1\times10^{-3}$. Each PPO iteration collects rollouts of 32 environment steps, followed by 5 learning epoch using 4 minibatches. We set the discount factory $\gamma=0.99$, GAE lambda $\lambda=0.95$, clipping ratio 0.2, and value loss coefficient 1.0. The reward definitions for both the recovery and locomotion tasks are detailed in Table~\ref{table:rewards}. Following the implementation of \cite{standup2}, the rewards for the recovery task are also grouped into four categories.

During the distillation process, we construct the  actor network as an MoE architecture to improve the policy's capacity. The MoE actor comprises 2 experts, each implemented as an MLP with the same hidden dimensions as the behavior-specific policy. The gate network, which determines the mixing weights of the experts' output actions, is also implemented as an MLP with hidden dimensions [512, 256, 128]. The total distillation loss is defined as a weighted sum of the two components:
\begin{equation}
\begin{aligned}
\mathcal{L}_{\text{distill}} =\; & \lambda_{\text{MSE}} \cdot \mathbb{E}_{a^d \sim \pi^d,a^b\sim\pi^b} \left[\|a^d - a^b\|^2_2\right] \\
& + \lambda_{\text{KL}} \cdot \mathbb{E}\left[\text{KL}\left(\pi^d \| \pi^b\right)\right]
\end{aligned}
\end{equation}
where the $\lambda_{\text{MSE}}$ and $\lambda_{\text{KL}}$ are set to 0.1 and 0.5 respectively. The procedure is summarized in Algorithm~\ref{alg:distillation}. In our implementation, the number of parallel environments $N$ is 4096, the rollout length $T$ is 32, the number of update epochs $K$ is 5, and the minibatch size $B$ is 4. The selection of $\pi^b$ based on $s_t$ is determined by the robot's base height: if the base height is greater than 0.5 $m$, the walk behavior policy is used; otherwise, the recovery behavior policy is selected. The student policy is updated using a learning rate of $1\times 10^{-3}$.

\begin{algorithm}[h]
\caption{Behavior Cloning via Multi-Expert Distillation}
\label{alg:distillation}
\begin{algorithmic}[1]
\REQUIRE Behavior-specific policies $\pi_{r}^b, \pi_{w}^b$, Multi-behavior policy $\pi^d$, number of environments $N$, rollout length $T$, number of update epochs $K$, minibatch size $B$
\STATE Initialize storage $\mathcal{D}$
\FOR{iteration = $1,2,\dots$}
    \STATE Collect rollouts in $N$ parallel environments:
    \FOR{t = $1$ to $T$}
        \STATE Observe current state $s_t$
        \STATE Select behavior policy $\pi^b$ based on $s_t$
        \STATE $a_t^b,\mu_t^b,\sigma_t^b \gets \pi^b(s_t)$\:\://get expert action
        \STATE $a_t^d,\mu_t^d,\sigma_t^d \gets \pi^d(s_t)$\:\://get student action
        \STATE Store $(a_t^b,\mu_t^b,\sigma_t^b,a_t^d,\mu_t^d,\sigma_t^d)$ in $\mathcal{D}$
    \ENDFOR
    \FOR{epoch = $1$ to $K$}
        \STATE Sample minibatches of size $B$ from $\mathcal{D}$
        \STATE Compute loss $\mathcal{L}$ (Eq.~(9))
        \STATE Update $\pi^d$ via gradient descent on $\mathcal{L}$
    \ENDFOR
    \STATE Clear storage $\mathcal{D}$
\ENDFOR
\end{algorithmic}
\end{algorithm}

\paragraph{RL Fine-tuned Policy}
% rl fine-tune: reward functions(same as behavior specific policy), network&hyper-params(same as distillation policy, while the lr is 1e-4)
We initialize the second-stage RL fine-tuning process using the multi-behaviors policy $\pi^d$. The network architecture remains unchanged, allowing the policy to retain its previously acquired multi-skill knowledge. During this stage, the policy is further fine-tuned on complex terrains, using the same task-specific reward structures as in the initial behavior-specific training (i.e., locomotion and recovery rewards described in Table~\ref{table:rewards}). In practice, we observed that using the same learning rate as in the first stage often led to training failures in the second stage. We reduce the policy learning rate to $1\times10^{-4}$, which mitigates the risk of policy collapse and excessive forgetting of previously acquired skills.

To alleviate gradient interference between different task objectives during fine-tuning, we employ a gradient surgery strategy following the procedure described in~\cite{yu2020gradient}. Specifically, after computing the task-specific policy gradients $\bm{g}_w$ and $\bm{g}_r$ from the walk and recovery objectives, respectively, we project one gradient onto the normal plane of the other whenever a conflicting direction is detected (i.e., when $\langle \bm{g}_w, \bm{g}_r \rangle < 0$). Since our framework involves only two task, we randomly select the projection direction at each update step, i.e., either projecting the locomotion gradient onto the recovery gradient or vice versa. This stochastic projection prevents bias toward any single task and maintains balanced optimization across behaviors.

Both training stages use the same domain randomization settings and joint PD gains ($K_p$ and $K_d$), as shown in Table~\ref{table:dr} and Table~\ref{table:pd} respectively. The domain randomization terms follow the configurations commonly adopted in previous works~\cite{rudin2022learning,standup1}, while the ranges of randomized parameters are further adjusted based on empirical performance observed in real-world experiments. The randomization parameters are resampled at the beginning of each episode to prevent overfitting to specific simulation dynamics and to encourage policy robustness across varying environmental and physical conditions.
\\
\\
\\
\\
\\
\\
\\
{\renewcommand{\arraystretch}{1.3}
\begin{center}
\captionof{table}{Reward Functions Used in Both Training Stages. The $f_{\text{tol}}$ adopts a Gaussian-style formulation, as detailed in \cite{standup2}.}
\vspace{-0.8em}
\label{table:rewards}
\resizebox{\columnwidth}{!}{
\begin{tabular}{l c r}
\toprule
\textbf{Term} & \textbf{Equation} & \textbf{Scale}\\
\hline
\noalign{\vskip 0.1cm}
\multicolumn{3}{c}{\textbf{Walking $\bm{r}^w$}} \\ 
\hline
Track lin. vel. & exp$\{ -\frac{||\bm{v}_{\text{lin}}^{\text{cmd}} - \bm{v}_{\text{lin}}||^2_2}{0.25} \}$  & $2.0$ \\
Track ang. vel. & exp$\{ -\frac{(\bm{\omega}_{\text{yaw}}^{\text{cmd}} - \bm{\omega}_{\text{yaw}})^2}{0.25} \}$ & $2.0$ \\
Joint acc. & $\|\ddot{\theta}\|_2^2$ & $-5\text{e}-7$ \\
Joint vel. & $\|\dot{\theta}\|_2^2$ & $-1\text{e}-3$ \\
Action rate & $\|\bm{a}_t - \bm{a}_{t-1}\|_2^2$ & $-0.03$ \\
Action smoothness & $\|\bm{a}_t - 2\bm{a}_{t-1}+\bm{a}_{t-2}\|_2^2$ & $-0.05$ \\
Angular vel. ($x y$)& $\|\bm{\omega}_{xy}\|_2^2$ & $-0.05$ \\
Orientation & $\|\bm{g}_{xy}\|_2^2$ & $-2.0$ \\
Joint power & $|\tau||\dot{ \theta}|^{T}$ & $-2.5\text{e}-5$ \\
Feet clearance & $\displaystyle\sum_{\text{foot}} ((z^{i}-h^{\text{target}})^2* \|\bm{v}^{i}_{xy}\|)$ & $-0.25$ \\
Feet stumble & $\mathbb{I}(\exists i, |\bm{F}_i^{xy}| \ge 3|F_i^z|)$ & $-1.0$ \\
Torques & $\displaystyle\sum_{\text{all joints}}\tau_i^2$ & $-1e-5$ \\
Arm joint deviations & $\displaystyle\sum_{\text{arm joints}}|\theta_i - \theta_{\text{default}}|$ & $-0.5$ \\
Hip joint deviations & $\displaystyle\sum_{\text{hip joints}}|\theta_i - \theta_{\text{default}}|$ & $-0.5$ \\
Joint pos. limits & $\displaystyle\sum_{\text{all joints}}\bm{out}_i$ & $-2.0$ \\
Joint vel. limits & $RELU(\dot \theta - \dot \theta^{\text{max}})$ & $-1.0$ \\
Torque limits & $RELU(\tau - \tau^{\text{max}})$ & $-1.0$ \\
% Feet lateral dist. & $|y^{\text{base}}_i-y^{\text{base}}_j| - d_{\text{min}}$ & $0.5$ \\
Feet slippage & $\displaystyle\sum_{\text{feet}}|\bm{v}_i^{\text{foot}}| * \mathbb{I}_{\text{contact}}$ & $-0.25$ \\
Collision & $n_{\text{collision}}$ & $-15.0$ \\
Feet air time & $\displaystyle\sum_{\text{foot}}(t^{\text{air}}_{i}-0.5)*\mathbb{I}(\text{first contact}_i)$ & $1.0$ \\
Stuck & $(\|\bm{v}\|_2\le0.1)*(\|\bm{c}^v\|_2 \ge 0.2)$ & $-1.0$ \\
\hline
\multicolumn{3}{c}{\textbf{Fail Recovery $\bm{r}^r$}} \\ 
\hline
\multicolumn{3}{c}{\textbf{Task Reward}} \\ 
Orientation & $f_{\text{tol}}(-\theta^z_{\text{base}},[0.99,\text{inf}],1,0.05)$ & $1.0$ \\
Head height & $f_{\text{tol}}(h_{\text{head}},[1,\text{inf}],1,0.1)$ & $1.0$ \\
\hline
\multicolumn{3}{c}{\textbf{Style Reward}} \\ 
Hip joint deviation & $\displaystyle\sum_{\text{hips}}\mathbb{I}(\text{max}|\theta_i|>0.9\vee\text{min}|\theta_i|<0.8)$ & $-10.0$ \\
Knee deviation & $\displaystyle\sum_{\text{knees}}\mathbb{I}(\text{max}|\theta_i|>2.85\vee\text{min}|\theta_i|<-0.06)$ & $-0.25$ \\
Shoulder roll deviation & $\mathbb{I}(\theta_\text{left}<-0.02\vee\theta_\text{right}>0.02)$ & $-2.5$ \\
Thigh orientation & $f_\text{tol}(\frac{1}{2}\displaystyle\sum_\text{thighs}(\theta^{\text{z}}_\text{thigh}),[0.8,\text{inf}],1,0.1)$ & $10.0$ \\
Feet distance & $\mathbb{I}(\|\bm{p}^{xy}_{\text{left\_feet}} - \bm{p}^{xy}_{\text{right\_feet}}\|^2>0.9)$ & $-10.0$ \\
Angular vel. ($x,y$) & $\text{exp}(-2\| \omega_{xy}\|^2_2)*\mathbb{I}(h_\text{base}>h_\text{stage1})$ & $25.0$ \\
% Foot displacement & $\text{exp}(\text{clip}(-2\| \bm{q}^{xy}_\text{base}-\bm{q}^{xy}_\text{foot} \|^2,0.3,\text{inf}))*\mathbb{I}(h_\text{base}>h_\text{stage2})$ & $2.5$ \\
Foot displacement & $\text{exp}(\text{clip}(-2\| \bm{q}^{xy}_\text{base}-\bm{q}^{xy}_\text{foot} \|^2,0.3,\text{inf}))$ & $2.5$ \\
\hline
\multicolumn{3}{c}{\textbf{Regularization Reward}} \\ 
Joint acc. & $\|\ddot{\theta}\|_2^2$ & $-2.5\text{e}-7$ \\
Joint vel. & $\|\dot{\theta}\|_2^2$ & $-1\text{e}-3$ \\
Action rate & $\|\bm{a}_t - \bm{a}_{t-1}\|_2^2$ & $-0.01$ \\
Action smoothness & $\|\bm{a}_t - 2\bm{a}_{t-1}+\bm{a}_{t-2}\|_2^2$ & $-0.05$ \\
Torques & $\displaystyle\sum_{\text{all joints}}\tau_i^2$ & $-1e-5$ \\
Joint power & $|\tau||\dot{ \theta}|^{T}$ & $-2.5\text{e}-5$ \\
Target joint track & $|\tau||\dot{ \theta}|^{T}$ & $-2.5\text{e}-5$ \\
Joint pos. limits & $\displaystyle\sum_{\text{all joints}}\bm{out}_i$ & $-2.0$ \\
Joint vel. limits & $RELU(\dot \theta - \dot \theta^{\text{max}})$ & $-1.0$ \\
\hline
\multicolumn{3}{c}{\textbf{Post-Task Reward}} \\ 
Angular vel. ($x,y$) & $\text{exp}(-2\| \bm{\omega}_{xy}\|_2^2)*\mathbb{I}(h_\text{base}>h_\text{stage3})$ & $10.0$ \\
Base lin. vel. ($x,y$) & $\text{exp}(-5\| \bm{v}_{xy}\|_2^2)*\mathbb{I}(h_\text{base}>h_\text{stage3})$ & $10.0$ \\
Orientation & $\text{exp}(-5\| \bm{g}_{xy}\|_2^2)*\mathbb{I}(h_\text{base}>h_\text{stage3})$ & $10.0$ \\
Base height & $\text{exp}(-20| h_\text{base}-0.75|)*\mathbb{I}(h_\text{base}>h_\text{stage3})$ & $10.0$ \\
Target joint deviations & $\text{exp}(-0.1\displaystyle\sum_{\text{all joints}}(\theta_i-\theta_i^{\text{default}})^2)$ & $10.0$ \\
Target feet distance & $f_\text{tol}(y_\text{left feet}-y_\text{right feet},[0.3,0.4],0.1,0.05)$ & $-5.0$ \\
\hline
\noalign{\vskip 0.1cm}
% \textbf{Style Reward $\bm{r}^\text{style}$} & $\displaystyle \text{max}[0, 1-\frac{1}{4}(D_{\phi}(\tau)-1)^2]$ & 5.0 (walking) \\
\multirow{2}{*}{\textbf{Style Reward $\bm{r}^\text{style}$}} 
& \multirow{2}{*}{$\displaystyle \text{max}\left[0, 1 - \frac{1}{4}(D_{\phi}(\tau) - 1)^2\right]$} 
& 5.0 (Walking) \\
& & 50.0 (Recovery) \\
\noalign{\vskip 0.1cm}
\bottomrule
\end{tabular}
}
\end{center}
}

{\renewcommand{\arraystretch}{1.2}
\begin{center}
\captionof{table}{Domain Randomization Settings and Ranges During Both Training Stages}
\label{table:dr}
\resizebox{\columnwidth}{!}{
\begin{tabular}{l l l}
\hline
\textbf{Term} & \textbf{Randomization Range} & \textbf{Unit}\\
\hline
Restitution & [0, 1] & - \\
Friction coefficient & [0.1, 1] & - \\
Base CoM offset & [-0.03, 0.03] & m \\
Mass payload & [-2, 5] & Kg \\
Link mass & [0.8, 1.2]$\times$ default value & Kg \\
$K_p$ Gains & [0.8, 1.25] & Nm/rad \\
$K_d$ Gains & [0.8, 1.25] & Nms/rad \\
Actuation offset & [-0.05, 0.05] & Nm \\
Motor strength & [0.8, 1.2]$\times$ motor torque & Nm \\
Actions delay & [0, 100] & ms \\
Initial joint angle scale & [0.85, 1.15]$\times$ default value & rad \\
Initial joint angle offset & [-0.1, 0.1] & rad \\
\hline
\end{tabular}}
\end{center}
}

\setlength{\tabcolsep}{18pt} 
\renewcommand{\arraystretch}{1.2}
\captionof{table}{PD gains (\(K_p\), \(K_d\)) used for each joint during training and deployment.}
\label{table:pd}
\vspace{0.5em}
\begin{center}
\begin{tabular}{l l l}
\hline
\textbf{Joint} & $\bm{K_p}$ & $\bm{K_d}$ \\
\hline
Hip & 150 & 4 \\
Knee & 200 & 6 \\
Ankle & 40 & 2 \\
Shoulder & 40 & 4 \\
Elbow & 100 & 4 \\
\hline
\end{tabular}
\end{center}

\end{document}